\documentclass{article}

\usepackage{amsmath,amsfonts,bm}

\def\eqref#1{equation~\ref{#1}}

\def\1{\bm{1}}

\DeclareMathAlphabet{\mathsfit}{\encodingdefault}{\sfdefault}{m}{sl}
\SetMathAlphabet{\mathsfit}{bold}{\encodingdefault}{\sfdefault}{bx}{n}

\newcommand{\R}{\mathbb{R}}

\newcommand{\KL}{D_{\mathrm{KL}}}

\usepackage{arxiv}

\usepackage[utf8]{inputenc} 
\usepackage[T1]{fontenc}    
\usepackage{hyperref}       
\usepackage{url}            
\usepackage{booktabs}       
\usepackage{amsfonts}       
\usepackage{nicefrac}       
\usepackage{microtype}      
\usepackage{xcolor}         

\usepackage{mathtools}
\usepackage{graphicx}
\usepackage{subfigure}
\usepackage{cuted}

\DeclarePairedDelimiter{\norm}{\lVert}{\rVert}
\DeclareMathOperator{\diag}{diag}
\renewcommand{\KL}[2]{\text{KL}(#1~\Vert~#2)}
\DeclareMathOperator{\ECR}{ECR}
\DeclareMathOperator{\JSD}{JSD}
\DeclareMathOperator{\RTE}{RTE}
\renewcommand{\R}{\mathbb{R}}
\renewcommand{\b}[1]{\textbf{#1}}

\title{Robust Temporal Ensembling for Learning with Noisy Labels}

\author{%
  Abel Brown \\
  NVIDIA \\
  \texttt{abelb@nvidia.com} \\
  \And
  Benedikt Schifferer \\
  NVIDIA \\
  \texttt{bschifferer@nvidia.com} \\
  \And
  Robert DiPietro \\
  NVIDIA \\
  \texttt{rdipietro@nvidia.com} \\
}

\begin{document}

\maketitle

\begin{abstract}
Successful training of deep neural networks with noisy labels is an essential capability as most real-world datasets contain some amount of mislabeled data.  Left unmitigated, label noise can sharply degrade typical supervised learning approaches.  In this paper, we present \emph{robust temporal ensembling} (RTE), which combines robust loss with semi-supervised regularization methods to achieve noise-robust learning.  We demonstrate that RTE achieves state-of-the-art performance across the CIFAR-10, CIFAR-100, ImageNet, WebVision, and Food-101N datasets, while forgoing the recent trend of label filtering and/or fixing. Finally, we show that RTE also retains competitive corruption robustness to unforeseen \emph{input} noise using CIFAR-10-C, obtaining a mean corruption error (mCE) of 13.50\% even in the presence of an 80\% noise ratio, versus 26.9\% mCE with standard methods on clean data.
\end{abstract}

\section{Introduction}

Deep neural networks have enjoyed considerable success across a variety of domains, and in particular computer vision, where the common theme is that more labeled training data yields improved model performance \cite{2017arXiv171200409H, 2018arXiv180500932M, 2019arXiv191104252X, 2019arXiv191211370K}.  However, performance depends on the quality of the training data, which is expensive to collect and inevitably imperfect. For example, ImageNet \cite{deng2009} is one of the most widely-used datasets in the field of deep learning and despite over 2 years of labor from more than 49,000 human annotators across 167 countries, it still contains erroneous and ambiguous labels \cite{li2017, karpathy2014}. It is therefore essential that learning algorithms in production workflows leverage noise robust methods.

Noise robust learning has a long history and takes many forms \cite{NIPS2013_5073,6685834,2020arXiv200708199S}.  Common strategies include loss correction and reweighting \cite{patrini2016,zhang2018,Menon2020Can}, label refurbishment \cite{reed2014,song2019}, abstention \cite{thulasidasan2019}, and relying on carefully constructed \emph{trusted} subsets of human-verified labeled data \cite{liiccv2017,NIPS2018_8246,zhang2020}.  Additionally, recent methods such as SELF \cite{nguyen2020} and DivideMix \cite{li2020} convert the problem of learning with noise into a semi-supervised learning approach by splitting the corrupted training set into clean labeled data and noisy unlabeled data at which point semi-supervised learning methods such as Mean Teacher \cite{tarvainen2017} and MixMatch \cite{48557} can be applied directly.  In essence, these methods effectively discard a majority of the label information so as to side-step having to learning with noise at all.  The problem here is that noisy label filtering tactics are imperfect resulting in corrupted data in the small labeled partition and valuable clean samples lost to the large pool of unlabeled data.  Moreover, caution is needed when applying semi-supervised methods where the labeled data is not sampled i.i.d. from the pool of unlabeled data \cite{NEURIPS2018_c1fea270}.  Indeed, filtering tactics can be biased and irregular, driven by specification error and the underlying noise process of the label corruption.  Recognizing the success of semi-supervised approaches, we ask: can we leverage the underlying mechanisms of semi-supervised learning such as entropy regularization for learning with noise without discarding our most valuable asset, the labels?

\section{Robust Temporal Ensembling}
\subsection{Preliminaries}
\label{sec:prelim}
Adopting the notation of \cite{zhang2018}, we consider the problem of classification where $\mathcal{X}\subset \mathbb{R}^d$ is the feature space and $\mathcal{Y} = \{1,\ldots,c\}$ is the label space where the classifier function is a deep neural network with a softmax output layer that maps input features to distributions over labels $f: \mathcal{X}\rightarrow \mathbb{R}^c$.  The dataset of training examples containing \emph{in-sample} noise is defined as $D = \{(x_i,\tilde{y}_i)\}^{n}_{i=1}$ where $(x_i,\tilde{y}_i) \in (\mathcal{X} \times \mathcal{Y})$ and $\tilde{y}_i$ is the noisy version of the true label $y_i$ such that $p(\tilde{y}_i=k\vert y_i = j,x_i) \equiv \eta_{ijk}$.  We do not consider \emph{open-set} noise \cite{wang2018}, in which there is a particular type of noise that occurs on inputs, $\tilde{x}$, rather than labels.  Following most prior work, we make the simplifying assumption that the noise is conditionally independent of the input, $x_i$, given the true labels. In this setting, we can write $\eta_{ijk}= p(\tilde{y}_i=k\vert y_i = j) \equiv \eta_{jk}$ which is, in general, considered to be \emph{class dependent} noise\footnote{See \cite{lee2019} for treatment of conditionally dependent \emph{semantic} noise such that $\eta_{ijk}\ne\eta_{jk}$.}$^\text{,}$\footnote{Note that \cite{patrini2016} define the noise transition matrix $T$ such that $T_{jk} \equiv \eta_{jk}$.}.

To aid in a simple and precise corruption procedure, we now depart from traditional notation and further decompose $\eta_{jk}$ as $p_j \cdot c_{jk}$, where $p_j \in [0,1]$ is the probability of corruption of the $j$-th class and $c_{jk} \in [0,1]$ is the relative probability that corrupted samples of class $j$ are labeled as class $k$, with $c_{i\neq j}\geq 0$, $c_{jj} = 0$ and $\sum_{k}{c_{jk}} = 1$.  A noisy dataset with $m$ classes can then be described as transition probabilities specified by
\begin{equation}
\label{eqn:F}
F = \diag(P) \cdot C + \diag(1-P) \cdot \mathcal{I}
\end{equation}
where $C \in \R^{m \times m}$ defines the system confusion or noise structure, $P \in \R^m$ defines the noise intensity or ratio for each class, and $\mathcal{I}$ is the identity matrix.  When $c_{jk} = c_{kj}$ the noise is said to be \emph{symmetric} and is considered \emph{asymmetric} otherwise. If ratio of noise is the same for all classes then $p_j = p$ and the dataset is said to exhibit \emph{uniform} noise.  For the case of uniform noise, equation (\ref{eqn:F}) interestingly takes the familiar form of the Google matrix equation \cite{RevModPhys.87.1261} as
\begin{equation}
\label{eqn:Fu}
F_p = p \cdot C + (1-p) \cdot \mathcal{I}
\end{equation}
Note that, by this definition, $\eta_{jj} = p \cdot c_{jj} = 0$ which prohibits $\tilde{y}_i = y_i$.  This ensures a true effective noise ratio of $p$.  For example, suppose there are $m=10$ classes and we wish to corrupt labels with 80\% probability. Then if corrupted labels are sampled from $\mathcal{Y}$ rather than $\mathcal{Y} \setminus \{y\}$, $\frac{1}{10} \cdot 0.8 = 8\%$ of the corrupted samples will not actually be corrupted, leading to a \emph{true} corruption rate of 72\%.  Therefore, despite prescribing $p=0.8$, the true effective noise ratio would be $0.72$, which in turn yields a $\frac{0.08}{1 - 0.8} = 40\%$ increase in clean labels, and this is indeed the case in many studies \cite{zhang2018,nguyen2020,li2020, zhang2020}.

\subsection{Methods}
\label{sec:methods}
Cross entropy based loss can achieve noise-robust properties by using a Box-Cox power transform to stabilize loss variance which can be shown to be a form of maximum likelihood estimation (MLE) \cite{ferrari2010}.  Additionally, pseudo-labeling \cite{10.2307/2285824} can be shown to be a form of entropy regularization \cite{Grandvalet2005} which in the framework of maximum a posterior (MAP) estimation encourages low-density separation between classes by minimizing the conditional entropy of the class probabilities of the noisy data \cite{Lee2013}.  That is, by minimizing entropy, the overlap of class probability distribution can be reduced.  The implicit assumption here is that classes are, in fact, well separated \cite{Chapelle2005}.  Moreover, MAP estimation itself acts as a regularization of MLE by incorporating a priori knowledge of related training examples in order to solve the ill-posed noisy learning objective and further prevent overfitting.  Indeed, entropy regularization is favorable in situations for which the joint distribution, $p(x,y)$, is mis-specified \cite{Grandvalet2005} which further underpins the motivation of pseudo-labeling as an apt basis for regularization. 

A noise-robust task loss is leveraged which can be seen as a generalization of mean absolute error (MAE) and categorical cross entropy (CCE) \cite{zhang2018}.  The idea is that CCE learns quickly, but more emphasis is put on difficult samples which is prone to overfit noisy labels, while MAE treats all samples equally, providing noise-robustness but learning slowly.  To exploit the benefits of both MAE and CCE, a negative Box-Cox transformation \cite{10.2307/2984418} is used to stabilize the loss variance as
\begin{equation}
\label{eqn:gce}
\mathcal{L}_q(f(x_i),y_i=j) = \frac{(1-f_j(x_i)^q)}{q} 
\end{equation}  
where $q \in (0,1]$, and $f_j$ denotes the $j$-th element of $f$.  Note that this loss becomes CCE for $\lim_{q \to 0}\mathcal{L}_q$ and becomes MAE/unhinged loss when $q=1$.

Consistency regularization works under the assumption that a model should output similar predictions given augmented versions of the same input. This regularization strategy is a common component of semi-supervised learning algorithms with the general form of $\norm{p_\theta(y\vert x_\text{aug1}) - p_\theta(y\vert x_\text{aug2})}$ where $p_\theta(y\vert x)$ is the predicted class distribution produced by the model having parameters $\theta$ for input $x$ \cite{7780854, sajjadi2016}.  We build upon numerous variations from semi-supervised learning \cite{laine2017, tarvainen2017, 48557, Berthelot2020ReMixMatch:} and leverage an ensemble consistency regularization (ECR) strategy as
\begin{equation}
\label{eqn:meanteacher}
\ECR = \frac{1}{\vert \mathcal{Y} \vert N^*}\sum_{i=1}^{N^*}\norm{p_{\theta'}(y\vert x) - p_\theta(y\vert \mathcal{A}(x))}
\end{equation}
where $x$ is the training example, $\mathcal{A}$ is stochastic augmentation function reevaluated for each term in the summation, $\theta'_{t} = \alpha \theta'_{t-1} + (1-\alpha)\theta_t$ is a temporal moving average of model weights used to generate pseudo-label targets, and inputs are pre-processed with standard random horizontal flip and crop.  In practice, this consists of initializing a copy of the initial model and maintaining an exponential moving average as training progresses.  Some methods directly average multiple label predictions together at each optimization step to form a single pseudo-label target \cite{48557,li2020} but we find pseudo-label target distributions generated by $\theta'$ to be better suited for the learning with noise problem due to the intrinsic ensemble nature of the weight averaging process over many optimization steps \cite{tarvainen2017}.  In semi-supervised learning techniques, it is common to leverage a large batch-size of unlabeled data for consistency regularization.  However, we found that modulating $N^*$, rather than the batch size of the consistency term, yields a monotonic increase in model performance consistent with related works \cite{Berthelot2020ReMixMatch:}.  Moreover, in semi-supervised learning, different batches are used for supervised and unsupervised loss terms but we find (see Section \ref{sec:ablation}) that for the case of learning with noise, batches synchronized with task loss term yields superior performance. 

The Jensen-Shannon consistency loss is used to enforce a flat response of the classifier by incentivizing the model to be stable, consistent, and insensitive across a diverse range of inputs \cite{7780854}. The Jensen-Shannon divergence (JSD) is minimized across distributions $p_\text{orig}$, $p_\text{aug1}$, and $p_\text{aug2}$ of the original sample $x_\text{orig}$ and its augmented variants $x_\text{aug1}$ and $x_\text{aug2}$ which can be understood to measure the average information that the sample reveals about the identity of its originating distribution \cite{hendrycks2020augmix}.  This JSD term is computed with $M = (p_\text{orig}+p_\text{aug1}+p_\text{aug2})/3$ and is then
\begin{equation}
\label{eqn:jsd}
\JSD = \frac{1}{3}\Big(\KL{p_\text{orig}}{M} + \KL{p_\text{aug1}}{M} + \KL{p_\text{aug2}}{M}\Big)
\end{equation}
where $\KL{p}{q}$ is Kullback–Leibler divergence from $q$ to $p$.  In this way, the JSD term improves the stability of training in the presence of noisy labels and heavy data augmentation with a modest contribution to final classifier test accuracy as shown in Table \ref{tab:ablation_loo}.

\subsection{Putting It All Together}
We unify the various components defined in sections \ref{sec:methods} together under a single parsimonious loss function at training defined as
\begin{equation}
\label{eqn:rteloss}
L_{\RTE} = \mathcal{L}_q + \lambda_{\JSD} \cdot \JSD + \lambda_{\ECR} \cdot \ECR
\end{equation}
where the JSD term is synchronized with ECR by computing the clean distribution using $p_{\theta'}$. Final performance is reported using $\theta '$.  In practice we find AugMix \cite{hendrycks2020augmix} to be most performant at high levels of label noise as AugMix layers together several stochastically sampled augmentation chains in a convex combination which mitigates input degradation but also generates highly diverse transformations.  Because the ECR loss term is based on the Mean Squared Error between the probability predictions, its depends on the number of classes of the dataset since the average is calculated by the squared error per class. As we sum GCE, JSD and ECR terms, the weights $\lambda_{JSD}$ and $\lambda_{ECR}$ are adjusted so that associated loss terms have similar magnitudes.

Here, data augmentation serves dual purpose as a generic regularizer to mitigate over-fitting of noisy labels \cite{zhang2018mixup} as well as provides additional information about the vicinity or neighborhood of the training examples which is formalized by Vicinal Risk Minimization \cite{NIPS2000_ba9a56ce}. These augmented examples can be seen as drawn from a vicinity distribution of the training examples to enlarge support of the training distribution such that samples in the vicinity share the same class but does not model the relation across examples of different classes \cite{zhang2018mixup}.  Therefore, data augmentations approximate samples of nearby elements of the data manifold where the difference, $\xi(x)=\mathcal{A}(x)-x$, approximates elements of its tangent space \cite{athiwaratkun2018there}.  In this way, the ECR term can loosely be seen as generating a set of stochastic differential constraints at each optimization step of the classification task loss.  While stronger augmentation can enrich the vicinity distribution, augmentation methods such as MixUp \cite{zhang2018mixup} and RandAugment \cite{DBLP:conf/cvpr/CubukZSL20} can overly degrade training examples and drift off the data manifold \cite{hendrycks2020augmix}.  When learning with noise, it is therefore essential to leverage an augmentation process rich in variety but which also preserve the image semantics and local statistics so as to minimize the additional strain on an already ill-posed noisy learning objective.  Consistent with this understanding, although RandAugment has been successfully leveraged in semi-supervised learning \cite{Berthelot2020ReMixMatch:,49534,2019arXiv190412848X}, our experiments with RandAugment proved unsuccessful for extreme levels of label noise (Table \ref{tab:ablation_loo}).

\section{Related Work}

Some methods for learning with noise attempt to improve noisy learning performance head-on by leveraging augmentation as a strong regularizer to mitigate memorization of corrupted labels \cite{zhang2018mixup} while others attempt to refurbish corrupted labels to control the accumulation of noise from mislabeled data \cite{song2019}.  A recent theme in learning with noisy labels has been to transform the learning with noise problem into a semi-supervised one by removing the labels of training data determined to be corrupted to form the requisite dichotomy of clean labeled data and a pool of unlabeled data \cite{nguyen2020,li2020}; then directly applying semi-supervised approaches such as MixMatch \cite{48557} and MeanTeacher \cite{tarvainen2017}.  Other methods go so far as to require trusted human verified data and combine re-weighting with re-labeling into a meta optimization approach \cite{zhang2020}.  

Semi-supervised learning algorithms have advanced considerably in recent years, making heavy use of both data augmentation and consistency regularization.  MixMatch \cite{48557} proposed a low-entropy label-guessing approach for augmented unlabeled data and mixes labeled and unlabeled data using MixUp.  In MixMatch, pseudo-label targets are formed by averaging label distributions produce by the model on samples drawn from the vicinity distribution ($\frac{1}{K}\sum_{K}p_{\theta}(y\vert \mathcal{A}(x))$).  However, this averaging requires artificial sharpening to generate low-entropy pseudo-labels.  From the MAP estimation perspective, sharpening does not add auxiliary a priori knowledge for the optimization step but rather prescribes a desirable property of the model generated label distribution.  Indeed, our experiments with the use of artificial label sharpening in RTE resulted in failed training at high levels of label noise and subsequent related work recognized that stronger augmentations can result in disparate predictions so their average may not generate meaningful targets \cite{Berthelot2020ReMixMatch:}.  ReMixMatch \cite{Berthelot2020ReMixMatch:} introduced augmentation anchoring which aims to minimize the entropy between label distributions produced by multiple weak and strong data augmentations of unlabeled data using a control theory augmentation approach.  While pseudo-label guessing and augmentation anchoring motivate the utility of multiple augmentations of the same data, our proposed ECR for learning with noise differs in the following important ways: ECR does not use distribution alignment for ``fairness'', distribution averaging, or label-sharpening; ECR forms pseudo-label targets using an exponential average of model weights and is batch-synchronized with the task loss term.  Finally, the recent work, FixMatch \cite{49534}, proposes a simplified semi-supervised approach where the consistency regularization term uses hard pseudo-labeling for low-entropy targets together with a filtering step to remove low-confidence unlabeled examples but does not leverage multiple strong augmentations.

\section{Experiments}
\label{sec:expt}
In this section we analyze the performance of RTE against various uniform noise configurations for both symmetric and asymmetric settings, and against real-world label noise.  For asymmetric noise, we test both the traditional configuration \cite{patrini2016}, typically reported by related works, and an additional configuration defined by (\ref{eqn:c_asym_rte}) which is in the spirit of \cite{lee2019}, where we define the asymmetric noise structure using the confusion matrix of a trained shallow network.  In all of these experiments, RTE outperforms existing methods.  Finally, we perform additional ablation studies to better understand the contribution and synergy of the terms in equation (\ref{eqn:rteloss}) as well as to probe the efficacy of ECR.

In our first set of experiments we consider the standard CIFAR-10, CIFAR-100, and ImageNet datasets \cite{Krizhevsky2009LearningML,deng2009}. CIFAR-10 and CIFAR-100 each contain 50,000 training and 10,000 test images of 10 and 100 classes, respectively; and ImageNet contains approximately 1,000,000 training images and 50,000 validation images of 1000 classes. Additionally, we test networks trained with noisy labels against unforeseen input corruptions using CIFAR-10-C \cite{hendrycks2019robustness} which was constructed by corrupting the original CIFAR-10 test set with a total of 15 noise, blur, weather, and digital corruptions under different severity levels and intensities.  Classifier performance is averaged across these corruption types and severity levels to yield \emph{mean corruption error} (mCE).  Since CIFAR-10-C is used to measure network behavior under data shift, these 15 corruptions are not included during the training procedure.  Here, CIFAR-10-C helps to establish a rigorous benchmark for image classifier robustness to better understand how models trained with noisy data might perform in safety-critical applications.

To mitigate the sensitivity of experimental results to empirical, and perhaps arbitrary, choices of hyperparameters, we present additional results that leverage Population Based Training (PBT) \cite{2017arXiv171109846J, 2019arXiv190201894L} which is a simple asynchronous optimisation algorithm that jointly optimize a population of models and their hyperparameters.  In particular, PBT discovers a per-epoch \emph{schedule} of hyperparameter settings rather than a static fixed configuration used over the entirety of training.  These PBT schedules, for example, allow task loss $\mathcal{L}_q$ to vary between CE and MAE loss dynamically during training and similarly the number of ECR terms $N^*$ can be modulated to realize a form of curriculum learning.  Moreover, for our purposes, PBT schedules also provide a form of quasi-ablation study, as optimization of the task-loss parameter $q$, the number of ECR terms $N^*$, and the ECR weight $\lambda_{\ECR}$ allows for the realization of a simplified loss which forgos these components if determined maximally beneficial.  We find, as in other studies, that this joint optimization of hyperparameter schedules typically results in faster wall-clock convergence and higher final performance. \cite{2019arXiv190505393H,2019arXiv190201894L}.

\subsection{Uniform Symmetric Noise}
\label{sec:expt:us}
\textbf{Training Setup.} Training details can be found in Section \ref{appendix:uniformsetup} of the Supplementary Material.

\textbf{Baselines}. To best interpret the effectiveness of RTE, we compare our results to many techniques for learning with noise (Table \ref{tab:uniformnoise}). A description of each baseline method can be found in Section \ref{appendix:baselines} in the Supplementary Material.  Only two of these references provide ImageNet results trained with label noise (Table \ref{tab:uniformnoiseimagenet}).

\begin{figure}
  \centering
  \includegraphics[width=2.5in]{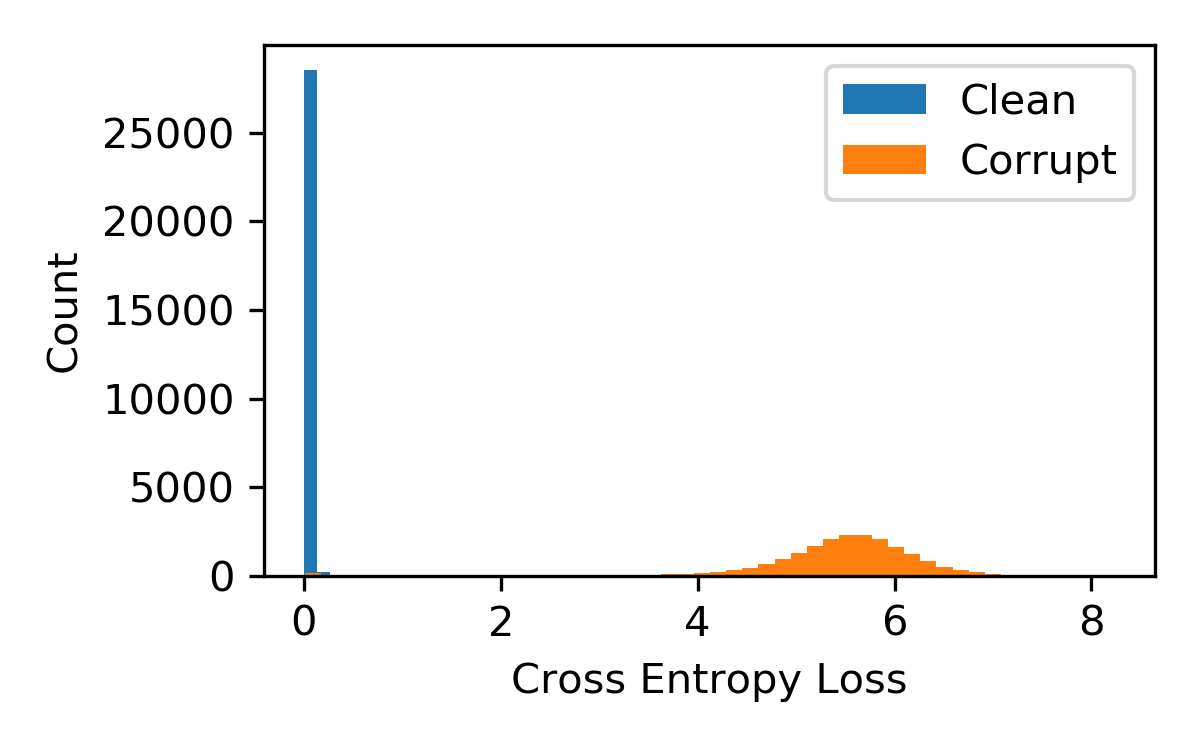}
  \includegraphics[width=2.5in]{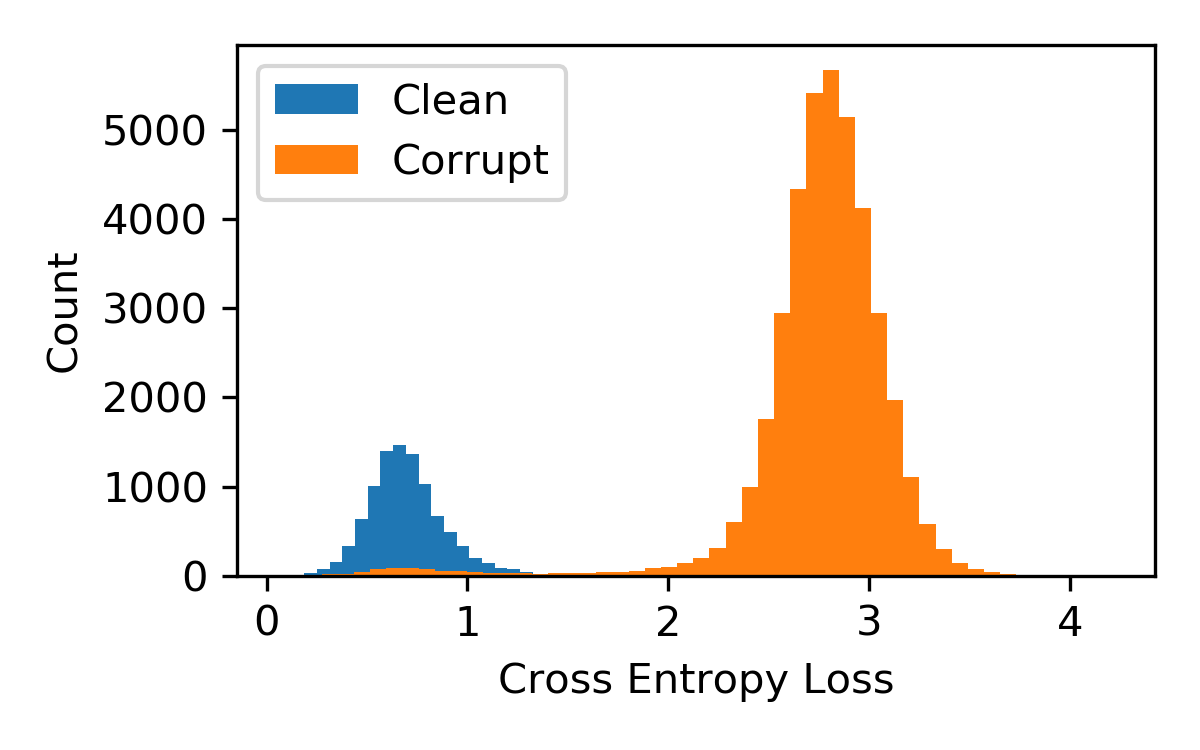}
  \caption{Loss distributions for clean labels versus corrupt labels on CIFAR-10 with 40\% label noise (left) and 80\% label noise (right). All losses are computed with respect to the labels used during training, which mimics a realistic setting (no access to clean labels).}
  \label{fig:loss-correct-vs-corrupt}
\end{figure}

\textbf{Results}. Experimental results with uniform symmetric noise for both CIFAR-10 and CIFAR-100 are presented in Table \ref{tab:uniformnoise} with comparisons to related work, including current state-of-the-art methods. RTE establishes new state-of-the-art performance at all noise levels and exhibits especially large performance gaps at high noise levels. At 80\% noise, previous state-of-the-art was achieved by \cite{arazo2019} in the case of CIFAR-10 and by \cite{li2020} in the case of CIFAR-100. RTE improves performance over these methods by 7.0 absolute percentage points and 6.2 absolute percentage points, respectively. Of all of these works, only two report results on ImageNet training with noisy labels. These are included alongside RTE results in Table \ref{tab:uniformnoiseimagenet}, where once again we see that RTE performs favorably, improving state-of-the-art performance in terms of both top-1 accuracy and top-5 accuracy. As in \cite{arazo2019} and \cite{li2020}, we also include loss distributions over clean and corrupt labels in Figure \ref{fig:loss-correct-vs-corrupt}. Here we can see that RTE prevents rote memorization of noisy labels. Moreover, Table \ref{tab:mce} shows that RTE retains strong corruption robustness with an mCE of 12.05\% and 13.50\% at noise ratios of 40\% and 80\% respectively, as measured using CIFAR-10-C.  Put in context, experiments summarized in Table \ref{tab:mce}  indicate that even with extreme levels of mislabeled training data, RTE trained models have lower corruption error than models trained using standard methods using clean data.
\begin{table*}[!htbp]
\renewcommand{\arraystretch}{1.3}
\centering
\caption{Test accuracy on CIFAR-10 and CIFAR-100 under uniform symmetric label noise. Results in parentheses are upper bounds since they were computed using lower noise levels (see sect. \ref{sec:prelim} for discussion). Note, our GCE-only results use true noise of 80\%, rather than the 72\% from the original GCE paper \cite{zhang2018}.  The results for Reed-Hard, S-Model \cite{goldberger2016}, Forward T and Co-Teaching are from \cite{nguyen2020} and the results for MixUp and Meta-Learning are from \cite{li2020}.  Finally, Polulation Based Training (PBT, see sect. \ref{sec:expt} for discussion) was used \emph{only} for RTE (PBT) experiments. That is, all non-CIFAR experiments, as well as the 'manual' CIFAR experiments, including baseline configurations, were performed \emph{without} PBT. The configurations for RTE (manual) and alternative configurations based on \cite{Berthelot2020ReMixMatch:} and \cite{hendrycks2020augmix}.  RTE provides better robustness to label noise than all other methods. Higher is better.}
\label{tab:uniformnoise}
\begin{tabular}{llllllll}
\toprule
Method                           &           &\multicolumn{3}{l}{CIFAR-10}    & \multicolumn{3}{l}{CIFAR-100}     \\
\hline
                                 & \# Params & \multicolumn{3}{l}{Noise Ratio} & \multicolumn{3}{l}{Noise Ratio} \\
                                 &           & 0\% & 40\%         & 80\%    &0\%      & 40\%           & 80\%        \\
\emph{(Prior Work)} \\[1ex]
Reed-Hard \cite{reed2014}       & --        && 69.66        & --          && 51.34        & --             \\
S-Model  \cite{goldberger2016}  & --        && 70.64        & --          && 49.10        & --             \\
MentorNet PD \cite{jiang2018}   & 84M       && 77           & 33          && 56           & 14           \\
Forward T \cite{patrini2016}    & --        && 83.25        & 54.64       && 31.05        & 8.90         \\
Open Set \cite{wang2018}        & --        && 78.15        & --          && --           & --             \\
Rand Weights \cite{ren2018}     & 36.4M     && 86.06        & --          && 58.01        & --             \\
Bi-Level \cite{jenni2018}       & 11.2M     && 89           & --          && 61.6         & --             \\
GCE \cite{zhang2018}            & 21.8M     && (87.12)      & (64.07)     && (61.77)      & (29.16)      \\
Co-Teaching \cite{han2018}      & --        && 81.85        & 29.22       && 55.95        & 23.22        \\
MixUp \cite{zhang2018mixup}     & --        && --           & (71.6)      && --           & (30.8)       \\
SELFIE \cite{song2019}          & --        && 86.5         & --          && 62.9         & --             \\
RoG \cite{lee2019}              & --        && 81.83        & --          && 55.68        & --             \\
M-DYR-H \cite{arazo2019}        & 11.2M     && --           & 86.6        && --           & 48.2         \\
PENCIL \cite{yi2019}            & 21.8M     && --           & --          && 69.12        & ``fail''         \\
Meta-Learning \cite{li2019}     & --        && --           & (77.4)      && --           & (42.4)       \\
SELF \cite{nguyen2020}          & 25.0M     && 93.70        & 69.91       && 71.98        & 42.09        \\
DivideMix \cite{li2020}         & 11.2M     && 94.9         & 79.8        && 75.2         & 60.2         \\[1ex]
\emph{(Our Work)} \\[1ex]
RTE (Manual)                     & 13.1M    & 95.67 & 94.84        & 93.09   & 79.71    & 76.70        & 64.02        \\
RTE (PBT)                        & 13.1M     && \b{95.52}    & \b{93.64}   && \b{77.44}    & \b{66.43}    \\
\hline
\emph{(Alternative Baseline Configurations)} \\[1ex]
CE-only                          &13.1M&& 90.06  & 59.66 && 65.98   & 35.80 \\
GCE-only                         &13.1M&& 91.35  & 59.15 && 69.73   & 39.19 \\
CE+JSD+ECR                       &13.1M&& 95.45	& 76.08 && 71.89   & 40.43 \\ 
\hline
\emph{(Alternative RTE Configuration)} \\[1ex]
RTE (PreAct ResNet-18 \cite{2016arXiv160305027H})       &11.2M&&& 92.00 &&&\\
\bottomrule
\end{tabular}
\end{table*}

\begin{table}
\renewcommand{\arraystretch}{1.3}
    \centering
    \caption{RTE mean corruption error on CIFAR-10-C for models trained at various uniform symmetric noise ratios.  Baseline reference values for `Standard' and `AugMix' mCE are reported from \cite{hendrycks2020augmix} using WRN 40x2 on clean data. Lower is better.}
    \begin{tabular}{lcc|ccc} 
        \toprule
        \multicolumn{3}{l}{} & \multicolumn{3}{l}{RTE with Noise Ratio:}\\
        &Standard & AugMix & 0\% & 40\% & 80\% \\
        \hline
        $\downarrow$ mCE & 26.9 & 11.2 & 11.5 & 12.05 & 13.50 \\
        \bottomrule
    \end{tabular}
    \label{tab:mce}
\end{table}

\begin{table}
\renewcommand{\arraystretch}{1.3}
    \centering
    \caption{Validation accuracy on ImageNet with 40\% uniform symmetric label noise. RTE hyperparameter configuration based on \cite{hendrycks2020augmix}.} 
    \begin{tabular}{lccc}
        \toprule%
                    & MentorNet \cite{jiang2018} & SELF \cite{nguyen2020}   & RTE \\
        \hline
        \# Params   & 59M       & 25.0M   & 25.6M     \\
        Top-1 Acc   & 65.1      & 71.31   & \b{74.79} \\
        Top-5 Acc   & 85.9      & 89.92   & \b{91.26} \\
        \bottomrule
    \end{tabular}
    \label{tab:uniformnoiseimagenet}
\end{table}

\subsection{Uniform Asymmetric Noise}
\label{sec:expt_asym}
\textbf{Training Setup}.  For consistency, uniform asymmetric noise experiments use the same hyperparameter configurations outlined for uniform symmetric noise. Here we test RTE performance using both the traditional asymmetric noise configuration \cite{patrini2016} typically reported by related works defined by Equation \ref{eqn:c_asym_patrini} in Section \ref{appendix:asym} of the Supplementary Material as well as an additional configuration in the spirit of \cite{lee2019} where we define the asymmetric noise structure using the confusion matrix of a trained shallow network defined by Equation \ref{eqn:c_asym_rte} in Section \ref{appendix:asymsetup} of the Supplementary Material.

The asymmetric noise defined by \cite{patrini2016} in equation (\ref{eqn:c_asym_patrini}) does not corrupt all classes but rather attempts to capture a noise process whereby labelers confuse specific pairs of classes which by some is argued to be more realistic in practice \cite{han2018,ren2018}.  We additionally consider a rich noise structure by training a shallow classifier (ResNet-10) on clean CIFAR-10 and use the resulting confusion matrix of this model to define the noise structure in equation (\ref{eqn:c_asym_rte}). For example, this asymmetric noise process readily captures the phenomenon that objects on blue backgrounds are often confused (e.g. birds, ships, and airplanes) and its natural asymmetry where $p(\tilde{y}_i=\tiny\textbf{SHIP}\vert y_i = \tiny\textbf{AIRPLANE})=$ 0.2772 while $p(\tilde{y}_i=\tiny\textbf{AIRPLANE} \vert y_i = \tiny\textbf{SHIP})=$ 0.4603 (locations $[1, 9]$ and $[9, 1]$ in Eq. \ref{eqn:c_asym_rte}).  Dataset statistics are provided for an instance of CIFAR-10 with asymmetric label noise prescribed according to equation (\ref{eqn:c_asym_rte}) with a uniform noise ratio of 60\% in Table \ref{tab:c_asym_sample_60} of Section \ref{sec:c_asym_sample_60} in the Supplementary Material.

\textbf{Baselines}. In the case of asymmetric noise as defined in \cite{patrini2016}, by equation (\ref{eqn:c_asym_patrini}), we compare the performance of RTE against existing work.  A brief description of each baseline method can be found in Section \ref{appendix:baselines} of the Supplementary Material.  In the case of asymmetric noise structure as defined in equation (\ref{eqn:c_asym_rte}), to our knowledge, prior work does not exist, and we report RTE performance at varied noise levels.

\textbf{Results}.
The results for asymmetric noise as presented in related works defined in \cite{patrini2016} by equation (\ref{eqn:c_asym_patrini}) with a uniform noise ratio of 40\% are shown in Table \ref{tab:asym_results_patrini} along side the performance of related methods. Again, RTE improves the state-of-the-art performance in this category, with a 1.1 absolute percentage point increase over \cite{li2020}.

Test accuracy for different level of asymmetric noise using $C$ defined by (\ref{eqn:c_asym_rte}) are shown in Table \ref{tab:asym_results_ours}. Even with 60\% noise ratio, RTE achieves 93.99\% test accuracy. The first significant decline in accuracy occurs around a 65\% asymmetric noise ratio, when the majority labels in a class are corrupted labels from another class. That is, for $F_{p=0.65}$ with $C$ defined by (\ref{eqn:c_asym_rte}), there are more \small{AUTOMOBILE}\normalsize \, images labeled as \small{TRUCK}\normalsize s, than actual \small{TRUCK}\normalsize \, images labeled as \small{TRUCK}\normalsize .

\begin{table}
\renewcommand{\arraystretch}{1.3}
    \centering
    \caption{Test accuracy on CIFAR-10 with asymmetric noise as defined in \cite{patrini2016} by equation (\ref{eqn:c_asym_patrini}).  Higher is better.}
    \begin{tabular}{lccccc}
        \toprule%
        & \multicolumn{4}{l}{Noise Ratio: 40\%} \\ 
        & GCE \cite{zhang2018} & SELF \cite{nguyen2020}  & PENCIL \cite{yi2019} & DivideMix \cite{li2020} & RTE \\
        \hline
        Acc    & 64.79 & 89.07 & 91.16 & 93.40 & \b{94.49}  \\
        \bottomrule
    \end{tabular}
    \label{tab:asym_results_patrini}
\end{table}

\begin{table}
\renewcommand{\arraystretch}{1.3}
    \centering
    \caption{RTE test performance on CIFAR-10 for different ratios of uniform asymmetric noise defined according to equation (\ref{eqn:c_asym_rte}).  Sharp declines in accuracy begin to occur at 65\% noise due to more \small{AUTOMOBILE}\normalsize \,  images labeled as \small{TRUCK}\normalsize, than actual \small{TRUCK}\normalsize \, images labeled as \small{TRUCK}\normalsize, and so on.}
    \begin{tabular}{lcccccc}
        \toprule
        & \multicolumn{6}{l}{Noise Ratio} \\ 
        & 20\% & 40\% & 60\% & 65\% & 70\% & 72\%\\
        \hline
        $\uparrow$ Acc & 95.34 & 94.82 & 93.99 & 80.55 & 72.12 & 59.70\\
        $\downarrow$ mCE    & 11.22 & 11.89 & 13.73 & 25.44 & 33.61 & 44.87 \\
        \bottomrule
    \end{tabular}
    \label{tab:asym_results_ours}
\end{table}


\subsection{Real-World Data with Noisy Labels}

\begin{table}
\renewcommand{\arraystretch}{1.3}
\centering
\caption{ImageNet validation accuracy when trained on WebVision. Prior results are from \cite{chen2019} and \cite{li2020}. Higher is better.}
\label{tab:webvision}
\begin{tabular}{lll}
\toprule
Method             & Top-1 Acc    & Top-5 Acc  \\
\hline
F-correction       & 57.36        & 82.36      \\
D2L                & 57.80        & 81.36      \\
MentorNet          & 57.80        & 79.92      \\
Decoupling         & 58.26        & 82.26      \\
Co-teaching        & 61.48        & 84.70      \\
Iterative-CF       & 61.60        & 84.98      \\
DivideMix          & 75.20        & 90.84      \\
RTE                & \b{80.84}    & \b{97.24}  \\
\bottomrule
\end{tabular}
\end{table}

Most prior work on learning with noisy labels focuses on synthetically added noise, as considered in the previous section. Here, we also consider two datasets with real-world label noise: WebVision \cite{li2017webvision} and Food-101N \cite{lee2018}. All experiments use ResNet-50. For WebVision, we follow the experimental setup in \cite{li2020}, which uses the first 50 classes that overlap with ImageNet. Hyperparameters for both datasets can be found in Table \ref{tab:manual_config} of the Supplementary Material. Results are shown alongside prior work in Tables \ref{tab:webvision} and \ref{tab:food101n}. RTE leads to state-of-the-art results in both cases, increasing top-1 accuracy from 77.32\% \cite{li2020} to 80.84\% when training on WebVision and evaluating on the ImageNet validation set (which is clean), and from 85.11\% to 86.46\% in the case of Food-101N.

\begin{table}
\renewcommand{\arraystretch}{1.3}
\centering
\caption{Test accuracy on Food-101N. All methods are based on the ResNet-50 architecture. The Baseline and CleanNet results are from \cite{lee2018}, and the Deep Self Learning result is from \cite{han2019}. Higher is better.}
\label{tab:food101n}
\begin{tabular}{ll}
\toprule
Method             & Top-1 Acc     \\
\hline
Baseline           & 81.44 \\
CleanNet (hard)    & 83.47 \\
CleanNet (soft)    & 83.95 \\
Deep Self Learning & 85.11 \\
RTE                & \b{86.46} \\
\bottomrule
\end{tabular}
\end{table}

\subsection{Ablation Studies}
\label{sec:ablation}
\begin{figure*}
  \centering
  \includegraphics[width=\textwidth]{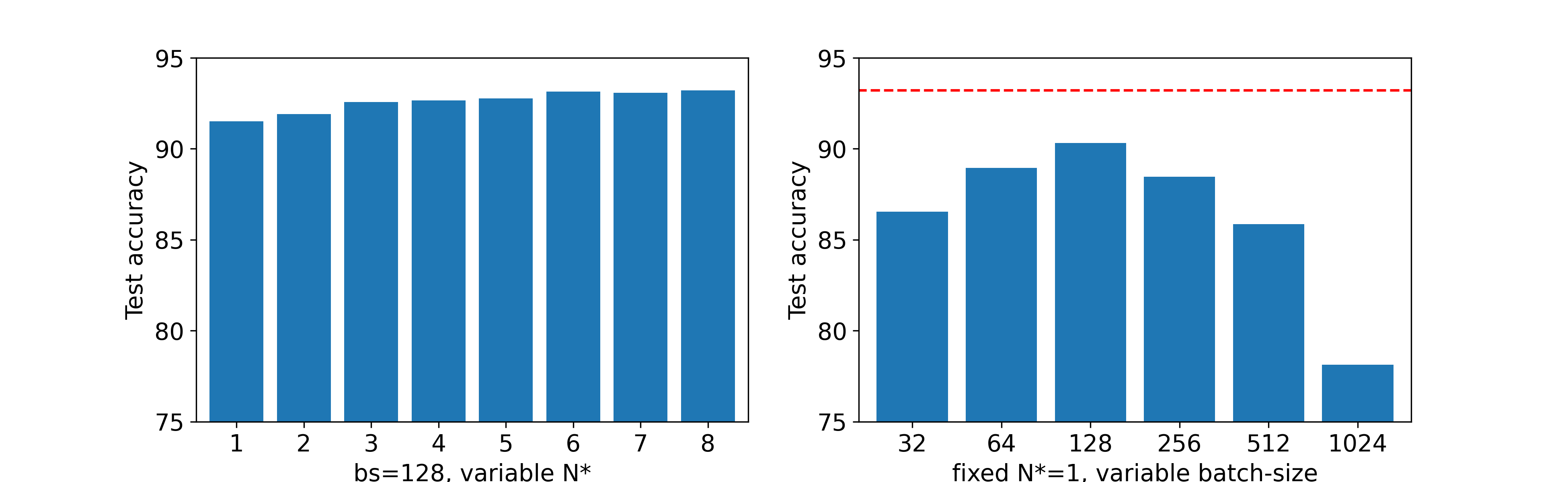}
  \caption{RTE ablation study using CIFAR-10 with uniform symmetric noise ratio of 80\%.  Left: the ECR batch entries are shared with the task loss and the batch size is fixed at 128, while the number of ECR terms ($N^*$) is varied.  Right: 1 ECR term is used with varying ECR batch size, using batch entries that are distinct from the task loss (analogous to a more traditional semi-supervised approach). The dashed red line on the right is the ECR baseline established using $N^* = 8$. }
  \label{fig:ecr_term_batch_size}
\end{figure*}

We perform various ablation studies to better understand the contribution of each term in equation (\ref{eqn:rteloss}), probe the efficacy of ECR, and compare with alternative regularization approaches.  Our ablation results are presented in Table \ref{tab:ablation_loo}.  These ablation studies use the training configurations defined in section \ref{sec:expt:us} unless otherwise stated.  First, because some prior work was carried out using a PreAct ResNet-18, e.g. DivideMix and M-DYR-H in Table \ref{tab:uniformnoise}, we provide results with the 28-layer Wide ResNet swapped out and a PreAct ResNet-18 swapped in. We can see that RTE's performance is minimally affected by this small difference in capacity: RTE achieves 93.09\% with a WRN and 92.00\% with a PreAct ResNet-18, vs. 79.8\% for DivideMix \cite{li2020} and 86.6\% for M-DYR-H \cite{arazo2019}, both using PreAct ResNet-18.  Next, we perform a component analysis where we remove one component at a time from equation \ref{eqn:rteloss} to better understand the performance contributions of each term. Removal of any term degrades performance.  We also test alternative consistency regularization approaches using label guessing as proposed in MixMatch \cite{48557} and augmentation anchoring from ReMixMatch \cite{Berthelot2020ReMixMatch:} which both underperform by significant margins compared to ECR.  Moreover, our results show significant benefits in the use of EMA whereas performance degrades with the augmentation anchoring approach consistent with prior work \cite{48557}.  Additionally, we test if label sharpening could benefit ECR, but we find that the artificial sharpening process amplifies noisy pseudo-labels early in training and learning collapses for high noise ratios.  Similarly, we find the strong linear chains of augmentations performed by RandAugment lead to training instabilities.  Figure \ref{fig:ecr_term_batch_size} summarizes the comparison of ECR to a traditional semi-supervised approach where a larger batch size is used for unsupervised regularization terms. This comparison indicates improved noisy learning performance with batch synchronization and repeated augmentation over larger batch sizes with single augmentations, validating the use of ECR for learning with noise.


\begin{table*}
\renewcommand{\arraystretch}{1.3}
\centering
\caption{Ablation study.  Test accuracy reported from CIFAR-10 with 80\% noisy labels.  Label guessing \cite{48557} and augmentation anchoring \cite{Berthelot2020ReMixMatch:} use a sharpening temperature of $T=0.5$ as recommended in the associated related works.}
\begin{tabular}{lclc}
\toprule
Ablation  & Test Acc & Ablation & Test Acc\\ \hline
RTE                               & 93.09 & Label Guessing, $K=2$          & 79.09  \\
RTE (PreAct ResNet-18) & 92.00 & Aug. Anchoring, $K=2$          & 83.59  \\
No ECR ($\lambda_{\ECR}=0$)       & 61.91 & Aug. Anchoring, $K=4$          & 83.24 \\
with CCE ($q=0$)                  & 76.08 & Aug. Anchoring, $K=6$          & 83.20  \\ 
No JSD ($\lambda_{\JSD}=0$)       & 90.37 & Aug. Anchoring, $K=2$, EMA     & 77.38 \\
with ECR, $N^*=2$, no EMA         & 67.23 & ECR with Label Sharpening      & fail\\
with ECR, $N^*=2$, no batch-sync  & 88.46 & ECR with RandAugment           & fail \\
with ECR, $N^*=2$, batch-sync     & 91.90 \\
\bottomrule
\end{tabular}
\label{tab:ablation_loo}
\end{table*}


\section{Conclusion}

We introduced robust temporal ensembling (RTE), which unifies semi-supervised regularization approaches and noise robust task loss as an effective method for learning with noisy labels. Rather than discarding noisy labels and applying semi-supervised methods, we successfully demonstrated a new approach for learning with noise which leverages all the data together without the need to filter, refurbish, or abstain from noisy training examples.  Through various experiments, we showed that RTE performs quite well in practice, advancing state-of-the-art performance across the CIFAR-10, CIFAR-100, and ImageNet datasets by 7.0, 6.2, and 3.5 absolute percentage points, respectively. Moreover, we demonstrated that RTE also performs well when training with data that exhibits \emph{real} label noise, achieving state-of-the-art results on the WebVision and Food-101N datasets. In addition, experiments summarized in Tables \ref{tab:asym_results_ours} and \ref{tab:mce} show that despite significant label noise, RTE trained models retain lower corruption error on unforeseen data shifts than models trained using standard methods on clean data. Finally, the results of numerous ablations summarized in section \ref{sec:ablation} validate the composition of loss terms and their combined efficacy over alternative methods. In future work, we are interested in the application of RTE for different modalities such as natural language processing and speech where label noise can be more pervasive and subjective.


\bibliographystyle{plain}
\bibliography{rte}

\begin{thebibliography}{10}

\bibitem{arazo2019}
Eric Arazo, Diego Ortego, Paul Albert, Noel~E O'Connor, and Kevin McGuinness.
\newblock Unsupervised label noise modeling and loss correction.
\newblock {\em arXiv preprint arXiv:1904.11238}, 2019.

\bibitem{athiwaratkun2018there}
Ben Athiwaratkun, Marc Finzi, Pavel Izmailov, and Andrew~Gordon Wilson.
\newblock There are many consistent explanations of unlabeled data: Why you
  should average.
\newblock In {\em International Conference on Learning Representations}, 2019.

\bibitem{Berthelot2020ReMixMatch:}
David Berthelot, Nicholas Carlini, Ekin~D. Cubuk, Alex Kurakin, Kihyuk Sohn,
  Han Zhang, and Colin Raffel.
\newblock Remixmatch: Semi-supervised learning with distribution matching and
  augmentation anchoring.
\newblock In {\em International Conference on Learning Representations}, 2020.

\bibitem{48557}
David Berthelot, Nicholas Carlini, Ian Goodfellow, Nicolas Papernot, Avital
  Oliver, and Colin Raffel.
\newblock Mixmatch: A holistic approach to semi-supervised learning.
\newblock In {\em NeurIPS}, 2019.

\bibitem{10.2307/2984418}
G.~E.~P. Box and D.~R. Cox.
\newblock An analysis of transformations.
\newblock {\em Journal of the Royal Statistical Society. Series B
  (Methodological)}, 26(2):211--252, 1964.

\bibitem{NIPS2000_ba9a56ce}
Olivier Chapelle, Jason Weston, L\'{e}on Bottou, and Vladimir Vapnik.
\newblock Vicinal risk minimization.
\newblock In T.~Leen, T.~Dietterich, and V.~Tresp, editors, {\em Advances in
  Neural Information Processing Systems}, volume~13, pages 416--422. MIT Press,
  2001.

\bibitem{Chapelle2005}
Olivier Chapelle and Alexander Zien.
\newblock Semi-supervised classification by low density separation.
\newblock {\em Proceedings of the Tenth International Workshop on Artificial
  Intelligence and Statistics, 57-64 (2005)}, 01 2005.

\bibitem{chen2019}
Pengfei Chen, Ben~Ben Liao, Guangyong Chen, and Shengyu Zhang.
\newblock Understanding and utilizing deep neural networks trained with noisy
  labels.
\newblock volume~97 of {\em Proceedings of Machine Learning Research}, pages
  1062--1070, Long Beach, California, USA, 09--15 Jun 2019. PMLR.

\bibitem{DBLP:conf/cvpr/CubukZSL20}
Ekin~D. Cubuk, Barret Zoph, Jonathon Shlens, and Quoc~V. Le.
\newblock Randaugment: Practical automated data augmentation with a reduced
  search space.
\newblock In {\em 2020 {IEEE/CVF} Conference on Computer Vision and Pattern
  Recognition, {CVPR} Workshops 2020, Seattle, WA, USA, June 14-19, 2020},
  pages 3008--3017. {IEEE}, 2020.

\bibitem{deng2009}
Jia Deng, Wei Dong, Richard Socher, Li-Jia Li, Kai Li, and Li~Fei-Fei.
\newblock Imagenet: A large-scale hierarchical image database.
\newblock In {\em 2009 IEEE conference on computer vision and pattern
  recognition}, pages 248--255, 2009.

\bibitem{RevModPhys.87.1261}
Leonardo Ermann, Klaus~M. Frahm, and Dima~L. Shepelyansky.
\newblock Google matrix analysis of directed networks.
\newblock {\em Rev. Mod. Phys.}, 87:1261--1310, Nov 2015.

\bibitem{li2017}
Li~Fei-Fei and Jia Deng.
\newblock Imagenet: Where have we been? where are we going?
\newblock
  \url{http://image-net.org/challenges/talks_2017/imagenet_ilsvrc2017_v1.0.pdf},
  2017.

\bibitem{ferrari2010}
Davide Ferrari and Yuhong Yang.
\newblock Maximum l q -likelihood estimation.
\newblock {\em Ann. Statist.}, 38(2):753--783, 04 2010.

\bibitem{6685834}
B.~{Frenay} and M.~{Verleysen}.
\newblock Classification in the presence of label noise: A survey.
\newblock {\em IEEE Transactions on Neural Networks and Learning Systems},
  25(5):845--869, 2014.

\bibitem{goldberger2016}
Jacob Goldberger and Ehud Ben-Reuven.
\newblock Training deep neural-networks using a noise adaptation layer.
\newblock 2016.

\bibitem{Grandvalet2005}
Yves Grandvalet and Y.~Bengio.
\newblock Semi-supervised learning by entropy minimization.
\newblock volume~17, 01 2004.

\bibitem{han2018}
Bo~Han, Quanming Yao, Xingrui Yu, Gang Niu, Miao Xu, Weihua Hu, Ivor Tsang, and
  Masashi Sugiyama.
\newblock Co-teaching: Robust training of deep neural networks with extremely
  noisy labels.
\newblock In S.~Bengio, H.~Wallach, H.~Larochelle, K.~Grauman, N.~Cesa-Bianchi,
  and R.~Garnett, editors, {\em Advances in Neural Information Processing
  Systems 31}, pages 8527--8537. Curran Associates, Inc., 2018.

\bibitem{han2019}
Jiangfan Han, Ping Luo, and Xiaogang Wang.
\newblock Deep self-learning from noisy labels.
\newblock In {\em Proceedings of the IEEE/CVF International Conference on
  Computer Vision}, pages 5138--5147, 2019.

\bibitem{He_2016_CVPR}
Kaiming He, Xiangyu Zhang, Shaoqing Ren, and Jian Sun.
\newblock Deep residual learning for image recognition.
\newblock In {\em Proceedings of the IEEE Conference on Computer Vision and
  Pattern Recognition (CVPR)}, June 2016.

\bibitem{2016arXiv160305027H}
Kaiming {He}, Xiangyu {Zhang}, Shaoqing {Ren}, and Jian {Sun}.
\newblock {Identity Mappings in Deep Residual Networks}.
\newblock {\em arXiv e-prints}, page arXiv:1603.05027, March 2016.

\bibitem{hendrycks2019robustness}
Dan Hendrycks and Thomas Dietterich.
\newblock Benchmarking neural network robustness to common corruptions and
  perturbations.
\newblock {\em Proceedings of the International Conference on Learning
  Representations}, 2019.

\bibitem{NIPS2018_8246}
Dan Hendrycks, Mantas Mazeika, Duncan Wilson, and Kevin Gimpel.
\newblock Using trusted data to train deep networks on labels corrupted by
  severe noise.
\newblock In S.~Bengio, H.~Wallach, H.~Larochelle, K.~Grauman, N.~Cesa-Bianchi,
  and R.~Garnett, editors, {\em Advances in Neural Information Processing
  Systems 31}, pages 10456--10465. Curran Associates, Inc., 2018.

\bibitem{hendrycks2020augmix}
Dan Hendrycks, Norman Mu, Ekin~Dogus Cubuk, Barret Zoph, Justin Gilmer, and
  Balaji Lakshminarayanan.
\newblock Augmix: A simple method to improve robustness and uncertainty under
  data shift.
\newblock In {\em International Conference on Learning Representations}, 2020.

\bibitem{2017arXiv171200409H}
Joel {Hestness}, Sharan {Narang}, Newsha {Ardalani}, Gregory {Diamos}, Heewoo
  {Jun}, Hassan {Kianinejad}, Md. Mostofa~Ali {Patwary}, Yang {Yang}, and Yanqi
  {Zhou}.
\newblock {Deep Learning Scaling is Predictable, Empirically}.
\newblock {\em arXiv e-prints}, page arXiv:1712.00409, December 2017.

\bibitem{2019arXiv190505393H}
Daniel {Ho}, Eric {Liang}, Ion {Stoica}, Pieter {Abbeel}, and Xi~{Chen}.
\newblock {Population Based Augmentation: Efficient Learning of Augmentation
  Policy Schedules}.
\newblock {\em arXiv e-prints}, page arXiv:1905.05393, May 2019.

\bibitem{2017arXiv171109846J}
Max {Jaderberg}, Valentin {Dalibard}, Simon {Osindero}, Wojciech~M.
  {Czarnecki}, Jeff {Donahue}, Ali {Razavi}, Oriol {Vinyals}, Tim {Green}, Iain
  {Dunning}, Karen {Simonyan}, Chrisantha {Fernando}, and Koray {Kavukcuoglu}.
\newblock {Population Based Training of Neural Networks}.
\newblock {\em arXiv e-prints}, page arXiv:1711.09846, November 2017.

\bibitem{jenni2018}
Simon Jenni and Paolo Favaro.
\newblock Deep bilevel learning.
\newblock In {\em Proceedings of the European Conference on Computer Vision
  (ECCV)}, pages 618--633, 2018.

\bibitem{jiang2018}
Lu~Jiang, Zhengyuan Zhou, Thomas Leung, Li-Jia Li, and Li~Fei-Fei.
\newblock Mentornet: Learning data-driven curriculum for very deep neural
  networks on corrupted labels.
\newblock In {\em International Conference on Machine Learning}, pages
  2304--2313, 2018.

\bibitem{karpathy2014}
Andrej Karpathy.
\newblock What i learned from competing against a convnet on imagenet.
\newblock
  \url{http://karpathy.github.io/2014/09/02/what-i-learned-from-competing-against-a-convnet-on-imagenet/},
  2014.

\bibitem{2019arXiv191211370K}
Alexander {Kolesnikov}, Lucas {Beyer}, Xiaohua {Zhai}, Joan {Puigcerver},
  Jessica {Yung}, Sylvain {Gelly}, and Neil {Houlsby}.
\newblock {Big Transfer (BiT): General Visual Representation Learning}.
\newblock {\em arXiv e-prints}, page arXiv:1912.11370, December 2019.

\bibitem{Krizhevsky2009LearningML}
A.~Krizhevsky.
\newblock Learning multiple layers of features from tiny images.
\newblock 2009.

\bibitem{49534}
Alex Kurakin, Chun-Liang Li, Colin Raffel, David Berthelot, Ekin~Dogus Cubuk,
  Han Zhang, Kihyuk Sohn, Nicholas Carlini, and Zizhao Zhang.
\newblock Fixmatch: Simplifying semi-supervised learning with consistency and
  confidence.
\newblock In {\em NeurIPS}, 2020.

\bibitem{laine2017}
Samuli Laine and Timo Alia.
\newblock Temporal ensembling for semi-supervised learning.
\newblock In {\em International Conference on Learning Representations}, 2017.

\bibitem{Lee2013}
Dong-Hyun Lee.
\newblock Pseudo-label : The simple and efficient semi-supervised learning
  method for deep neural networks.
\newblock {\em ICML 2013 Workshop : Challenges in Representation Learning
  (WREPL)}, 07 2013.

\bibitem{lee2019}
Kimin Lee, Sukmin Yun, Kibok Lee, Honglak Lee, Bo~Li, and Jinwoo Shin.
\newblock Robust inference via generative classifiers for handling noisy
  labels.
\newblock volume~97 of {\em Proceedings of Machine Learning Research}, pages
  3763--3772, Long Beach, California, USA, 09--15 Jun 2019. PMLR.

\bibitem{lee2018}
Kuang-Huei Lee, Xiaodong He, Lei Zhang, and Linjun Yang.
\newblock Cleannet: Transfer learning for scalable image classifier training
  with label noise.
\newblock In {\em Proceedings of the IEEE Conference on Computer Vision and
  Pattern Recognition}, pages 5447--5456, 2018.

\bibitem{2019arXiv190201894L}
Ang {Li}, Ola {Spyra}, Sagi {Perel}, Valentin {Dalibard}, Max {Jaderberg},
  Chenjie {Gu}, David {Budden}, Tim {Harley}, and Pramod {Gupta}.
\newblock {A Generalized Framework for Population Based Training}.
\newblock {\em arXiv e-prints}, page arXiv:1902.01894, February 2019.

\bibitem{li2020}
Junnan Li, Richard Socher, and Steven~C.H. Hoi.
\newblock Dividemix: Learning with noisy labels as semi-supervised learning.
\newblock In {\em International Conference on Learning Representations}, 2020.

\bibitem{li2019}
Junnan Li, Yongkang Wong, Qi~Zhao, and Mohan~S Kankanhalli.
\newblock Learning to learn from noisy labeled data.
\newblock In {\em Proceedings of the IEEE Conference on Computer Vision and
  Pattern Recognition}, pages 5051--5059, 2019.

\bibitem{li2017webvision}
Wen Li, Limin Wang, Wei Li, Eirikur Agustsson, and Luc Van~Gool.
\newblock Webvision database: Visual learning and understanding from web data.
\newblock {\em Preprint. arXiv}, 1708, 2017.

\bibitem{liiccv2017}
Yuncheng Li, Jianchao Yang, Yale Song, Liangliang Cao, Jiebo Luo, and Li-Jia
  Li.
\newblock Learning from noisy labels with distillation.
\newblock pages 1928--1936, 10 2017.

\bibitem{sgdr2017:}
Ilya Loshchilov and Frank Hutter.
\newblock Sgdr: Stochastic gradient descent with warm restarts.
\newblock In {\em International Conference on Learning Representations}, 2017.

\bibitem{2018arXiv180500932M}
Dhruv {Mahajan}, Ross {Girshick}, Vignesh {Ramanathan}, Kaiming {He}, Manohar
  {Paluri}, Yixuan {Li}, Ashwin {Bharambe}, and Laurens {van der Maaten}.
\newblock {Exploring the Limits of Weakly Supervised Pretraining}.
\newblock {\em arXiv e-prints}, page arXiv:1805.00932, May 2018.

\bibitem{10.2307/2285824}
G.~J. McLachlan.
\newblock Iterative reclassification procedure for constructing an
  asymptotically optimal rule of allocation in discriminant analysis.
\newblock {\em Journal of the American Statistical Association},
  70(350):365--369, 1975.

\bibitem{Menon2020Can}
Aditya~Krishna Menon, Ankit~Singh Rawat, Sashank~J. Reddi, and Sanjiv Kumar.
\newblock Can gradient clipping mitigate label noise?
\newblock In {\em International Conference on Learning Representations}, 2020.

\bibitem{NIPS2013_5073}
Nagarajan Natarajan, Inderjit~S Dhillon, Pradeep~K Ravikumar, and Ambuj Tewari.
\newblock Learning with noisy labels.
\newblock In C.~J.~C. Burges, L.~Bottou, M.~Welling, Z.~Ghahramani, and K.~Q.
  Weinberger, editors, {\em Advances in Neural Information Processing Systems
  26}, pages 1196--1204. Curran Associates, Inc., 2013.

\bibitem{nguyen2020}
Duc~Tam Nguyen, Chaithanya~Kumar Mummadi, Thi Phuong~Nhung Ngo, Thi Hoai~Phuong
  Nguyen, Laura Beggel, and Thomas Brox.
\newblock Self: Learning to filter noisy labels with self-ensembling.
\newblock In {\em International Conference on Learning Representations}, 2020.

\bibitem{NEURIPS2018_c1fea270}
Avital Oliver, Augustus Odena, Colin~A Raffel, Ekin~Dogus Cubuk, and Ian
  Goodfellow.
\newblock Realistic evaluation of deep semi-supervised learning algorithms.
\newblock In S.~Bengio, H.~Wallach, H.~Larochelle, K.~Grauman, N.~Cesa-Bianchi,
  and R.~Garnett, editors, {\em Advances in Neural Information Processing
  Systems}, volume~31, pages 3235--3246. Curran Associates, Inc., 2018.

\bibitem{patrini2016}
Giorgio {Patrini}, Alessandro {Rozza}, Aditya {Menon}, Richard {Nock}, and
  Lizhen {Qu}.
\newblock {Making Deep Neural Networks Robust to Label Noise: a Loss Correction
  Approach}.
\newblock {\em arXiv e-prints}, page arXiv:1609.03683, September 2016.

\bibitem{reed2014}
Scott {Reed}, Honglak {Lee}, Dragomir {Anguelov}, Christian {Szegedy}, Dumitru
  {Erhan}, and Andrew {Rabinovich}.
\newblock {Training Deep Neural Networks on Noisy Labels with Bootstrapping}.
\newblock {\em arXiv e-prints}, page arXiv:1412.6596, December 2014.

\bibitem{ren2018}
Mengye Ren, Wenyuan Zeng, Bin Yang, and Raquel Urtasun.
\newblock Learning to reweight examples for robust deep learning.
\newblock {\em arXiv preprint arXiv:1803.09050}, 2018.

\bibitem{sajjadi2016}
Mehdi Sajjadi, Mehran Javanmardi, and Tolga Tasdizen.
\newblock Regularization with stochastic transformations and perturbations for
  deep semi-supervised learning.
\newblock In D.~D. Lee, M.~Sugiyama, U.~V. Luxburg, I.~Guyon, and R.~Garnett,
  editors, {\em Advances in Neural Information Processing Systems 29}, pages
  1163--1171. Curran Associates, Inc., 2016.

\bibitem{song2019}
Hwanjun Song, Minseok Kim, and Jae-Gil Lee.
\newblock {SELFIE}: Refurbishing unclean samples for robust deep learning.
\newblock In Kamalika Chaudhuri and Ruslan Salakhutdinov, editors, {\em
  Proceedings of the 36th International Conference on Machine Learning},
  volume~97 of {\em Proceedings of Machine Learning Research}, pages
  5907--5915, Long Beach, California, USA, 09--15 Jun 2019. PMLR.

\bibitem{2020arXiv200708199S}
Hwanjun {Song}, Minseok {Kim}, Dongmin {Park}, and Jae-Gil {Lee}.
\newblock {Learning from Noisy Labels with Deep Neural Networks: A Survey}.
\newblock {\em arXiv e-prints}, page arXiv:2007.08199, July 2020.

\bibitem{JMLR:v15:srivastava14a}
Nitish Srivastava, Geoffrey Hinton, Alex Krizhevsky, Ilya Sutskever, and Ruslan
  Salakhutdinov.
\newblock Dropout: A simple way to prevent neural networks from overfitting.
\newblock {\em Journal of Machine Learning Research}, 15(56):1929--1958, 2014.

\bibitem{nesterov}
I.~Sutskever, J.~Martens, G.~Dahl, and G.~Hinton.
\newblock On the importance of initialization and momentum in deep learning.
\newblock {\em 30th International Conference on Machine Learning, ICML 2013},
  pages 1139--1147, 01 2013.

\bibitem{tarvainen2017}
Antti Tarvainen and Harri Valpola.
\newblock Mean teachers are better role models: Weight-averaged consistency
  targets improve semi-supervised deep learning results.
\newblock In I.~Guyon, U.~V. Luxburg, S.~Bengio, H.~Wallach, R.~Fergus,
  S.~Vishwanathan, and R.~Garnett, editors, {\em Advances in Neural Information
  Processing Systems 30}, pages 1195--1204. Curran Associates, Inc., 2017.

\bibitem{thulasidasan2019}
Sunil Thulasidasan, Tanmoy Bhattacharya, Jeff Bilmes, Gopinath Chennupati, and
  Jamal Mohd-Yusof.
\newblock Combating label noise in deep learning using abstention.
\newblock {\em arXiv preprint arXiv:1905.10964}, 2019.

\bibitem{wang2018}
Yisen Wang, Weiyang Liu, Xingjun Ma, James Bailey, Hongyuan Zha, Le~Song, and
  Shu-Tao Xia.
\newblock Iterative learning with open-set noisy labels.
\newblock In {\em Proceedings of the IEEE Conference on Computer Vision and
  Pattern Recognition (CVPR)}, June 2018.

\bibitem{2019arXiv190412848X}
Qizhe {Xie}, Zihang {Dai}, Eduard {Hovy}, Minh-Thang {Luong}, and Quoc~V. {Le}.
\newblock {Unsupervised Data Augmentation for Consistency Training}.
\newblock {\em arXiv e-prints}, page arXiv:1904.12848, April 2019.

\bibitem{2019arXiv191104252X}
Qizhe {Xie}, Minh-Thang {Luong}, Eduard {Hovy}, and Quoc~V. {Le}.
\newblock {Self-training with Noisy Student improves ImageNet classification}.
\newblock {\em arXiv e-prints}, page arXiv:1911.04252, November 2019.

\bibitem{yi2019}
Kun Yi and Jianxin Wu.
\newblock Probabilistic end-to-end noise correction for learning with noisy
  labels.
\newblock In {\em Proceedings of the IEEE Conference on Computer Vision and
  Pattern Recognition}, pages 7017--7025, 2019.

\bibitem{2016arXiv160507146Z}
Sergey {Zagoruyko} and Nikos {Komodakis}.
\newblock {Wide Residual Networks}.
\newblock {\em arXiv e-prints}, page arXiv:1605.07146, May 2016.

\bibitem{zhang2018mixup}
Hongyi Zhang, Moustapha Cisse, Yann~N. Dauphin, and David Lopez-Paz.
\newblock mixup: Beyond empirical risk minimization.
\newblock In {\em International Conference on Learning Representations}, 2018.

\bibitem{zhang2018}
Zhilu Zhang and Mert Sabuncu.
\newblock Generalized cross entropy loss for training deep neural networks with
  noisy labels.
\newblock In S.~Bengio, H.~Wallach, H.~Larochelle, K.~Grauman, N.~Cesa-Bianchi,
  and R.~Garnett, editors, {\em Advances in Neural Information Processing
  Systems 31}, pages 8778--8788. Curran Associates, Inc., 2018.

\bibitem{zhang2020}
Zizhao Zhang, Han Zhang, Sercan~O. Arik, Honglak Lee, and Tomas Pfister.
\newblock Distilling effective supervision from severe label noise.
\newblock In {\em Proceedings of the IEEE/CVF Conference on Computer Vision and
  Pattern Recognition (CVPR)}, June 2020.

\bibitem{7780854}
S.~{Zheng}, Y.~{Song}, T.~{Leung}, and I.~{Goodfellow}.
\newblock Improving the robustness of deep neural networks via stability
  training.
\newblock In {\em 2016 IEEE Conference on Computer Vision and Pattern
  Recognition (CVPR)}, pages 4480--4488, 2016.

\end{thebibliography}

\newpage

\appendix
\newpage

\section{Hyperparameters}

We include manual hyperparameter configurations in Table \ref{tab:manual_config}.   The configurations for CIFAR-10 and CIFAR-100 are based on \cite{Berthelot2020ReMixMatch:} and \cite{hendrycks2020augmix}. Experiment configurations were based on \cite{hendrycks2020augmix} for ImageNet, \cite{li2020} for Webvision, and \cite{han2019} for Food101N to facilitate comparison of results.

\begin{table*}
    \renewcommand{\arraystretch}{1.5}
    \centering
    \caption{Manual hyperparameter configurations. The two $\lambda$ values are $\lambda_{\JSD}$ and $\lambda_{\ECR}$. For WebVision and Food-101N, the base $b$ of the LR decay is computed such that the learning rate has decayed by 3 orders of magnitude at the end of training.}
    \begin{tabular}{lccccccccc}
    \\
    \toprule
              & Wt. Decay & $\beta$ & Dropout & BS  & LR                                & $\lambda$ & $q$                           & $N^*$ & $\alpha$ \\
    \hline\\[-2ex]
    CIFAR-10  & .001      & .9      & .01     & 128 & $\cos( \frac{7 \pi k}{16 K})$     & 12, 1     & $\sin( \frac{13\pi k}{16 K})$ & 10    & .99 \\
    CIFAR-100 & .0005     & .9      & .01     & 128 & 0.04                              & 5, 3      & 0.3                           & 8     & .99 \\
    ImageNet  & .001      & .9      & .00     & 256 & $10^{-1, -2, -3}$                 & 12, 10    & 0.3                           & 3     & .99 \\
    WebVision & .0001     & .9      & .00     & 256 & $10^{-3} \cdot b^k$                              & 12, 5     & 0.3                           & 3     & .99 \\
    Food-101N & 0.01      & .9      & .00     & 128 & $10^{-1, -2, -3}$                 & 12, 1     & 0.1                           & 2     & .99 \\
    \bottomrule
    \end{tabular}
    \label{tab:manual_config}
\end{table*}

\section{Baselines for Table \ref{tab:uniformnoise}}
\label{appendix:baselines}
In this section we provide a brief summary of the baseline methods in the main text:

\b{\cite{reed2014}} introduce two methods for achieving prediction consistency, one based on reconstruction and one based on bootstrapping, and demonstrated empirically that bootstrapping leads to better robustness to label noise.
\b{\cite{goldberger2016}} model the correct label as latent and having gone through a parameterized corruption process. Expectation maximization is used to estimate both the parameters of the corruption process and the underlying latent label.
\b{\cite{jiang2018}} introduce the idea of \emph{learning} a curriculum-learning strategy with a \emph{mentor} model to train a \emph{student} model to be robust to label noise.
\b{\cite{patrini2016}} estimate the noise transition matrix (under the assumption of feature independent noise) and show that, given the \emph{true} noise transition matrix, optimizing for the true underlying labels is possible.
\b{\cite{wang2018}} introduce an iterative scheme that combines 1. outlier detection in feature space (acting as a proxy to noisy-label detection), 2. a Siamese network (taking either a clean, clean pair or a clean, noisy pair) to encourage separation, and 3. sample reweighting based on clean vs. noisy confidence levels in order to effectively filter out noisy labels during training. They focus primarily on \emph{open-set} noise, but they also report performance of their system when used in the \emph{closed-set} setting.
\b{\cite{ren2018}} use a meta-learning approach to dynamically weight examples to minimize loss \emph{using a set of validation examples with clean labels}, however they also report a competitive baseline using a randomized weighting scheme which requires no clean validation set.
\b{\cite{jenni2018}} formulate example weighting as a bilevel-optimization problem, in which performance on a validation set is maximized with respect to example weights, subject to the constraint that the model maximizes performance on the training set; and they argue that this approach should lead to better generalization when label noise is present. 
\b{\cite{zhang2018}} introduce a loss function that is a generalization of cross-entropy loss and mean absolute error, which is beneficial since each exhibits distinct desirable properties: cross-entropy exhibits better gradient properties for learning, while mean absolute error exhibits better theoretically-grounded robustness to noisy labels. 
\b{\cite{han2018}} leverage co-teaching such that two networks are trained together, in which each network 1. identifies high-confidence examples, 2. passes this information in a message to its peer, and 3. leverages the incoming message to optimize using the examples selected by its peer.
\b{\cite{zhang2018mixup}} train using convex combinations of both input images and their labels, arguing that this approach makes it more difficult for the network to memorize corrupt labels.
\b{\cite{song2019}} measure label consistency throughout training in order to determine which samples are `refurbishable', and these samples are then `corrected' by replacing their ground-truth label with the most frequently-predicted label.
\b{\cite{lee2019}} do not modify the training process of the underlying neural network but instead form a generative model over the final (pre-softmax) features of the neural network, and this generative distribution along with Bayes rule is then used to estimate a more robust conditional distribution over the label.
\b{\cite{arazo2019}} fit a beta mixture model over the \emph{loss} using two mixture components, representing \emph{clean} and \emph{noisy} labels, and each sample's underlying component probabilities are used to weight each sample's contribution during training. They combine this approach with MixUp \cite{zhang2018mixup}.
\b{\cite{yi2019}} maintain a direct estimate of a distribution over true underlying labels during training, and train the parameters of a neural network by minimizing reverse KL divergence (from the model's predicted distribution to these true-label estimates). Meanwhile a `compatibility loss' is introduced to ensure that the estimated label distribution stays close to the noisy labels provided with the training set.
\b{\cite{li2019}} subject a student model to artificial label noise during training and take alternating gradient steps and maintain a teacher model that is not subjected to such noise. Here, alternating gradient steps are taken to 1. minimize classification loss and 2. minimize the KL divergence from the student's predicted distributions to the teacher's predicted distributions.
\b{\cite{nguyen2020}} use discrepancy between an ensemble-based teacher model and labels to identify and filter out incorrect labels, and continue to leverage these samples without the labels. This is done in a semi-supervised fashion by maintaining consistency between the teacher's predictions and the student's predictions.
\b{\cite{li2020}} maintain two networks and for each network models \emph{loss} using a mixture of Gaussians with two components (\emph{clean} and \emph{noisy}). Each network estimates which samples belong to each component, and the \emph{other} network then uses the \emph{clean} samples in a supervised manner along with the \emph{noisy} labels in a semi-supervised manner.

\section{Uniform Symmetric Noise Experimental Setup}
\label{appendix:uniformsetup}

For CIFAR-10, we leverage equation (\ref{eqn:Fu}) with $C^{10}_{i\neq j} = \frac{1}{9}$ and we employ a 28-layer residual network \cite{He_2016_CVPR} with a widening factor of 6 (WNR 28x6) \cite{2016arXiv160507146Z}, a dropout rate of 0.01 \cite{JMLR:v15:srivastava14a}, $\alpha = 0.99$, AugMix with a mixture width and severity of 3, a batch size of 128, and 300 epochs of training. We optimize using SGD with Nesterov momentum of 0.9 \cite{nesterov}, a weight decay of 0.001, and a cosine learning rate \cite{sgdr2017:} of $0.03 \cdot \cos(7\pi k / 16K)$, where $k$ is the current training step and $K$ is the total number of training steps. The RTE loss function (\ref{eqn:rteloss}) is configured with static $\lambda_{\JSD}$, $\lambda_{\ECR}$ and $N^*$ of 12, 1, and 10, respectively, whereas $q$ is scheduled according to $0.6\cdot \sin(13\pi k/16K)$ (which assigns small $q$-values in early training epochs, reaches a maximum of $q=0.6$ after 180 epochs, and decreases to $q=0.33$ over the remaining 120 epochs). For CIFAR-100, the setup is similar, but different hyperparameters are used; details are included in the Appendix in Table \ref{tab:manual_config}. In addition to manual configurations, we consider PBT with a population size of 35 to optimize learning rate, weight decay, $q$, $\lambda_{\JSD}$, $\lambda_{\ECR}$ and $N^*$.  Fastidious readers will find the complete PBT configuration defined in Appendix \ref{PBThyper}. For ImageNet, ResNet50 is used and trained with SGD for 300 epochs with a stepped learning rate of 0.1, 0.01 and 0.001 which begin at epochs 0, 100 and 200 respectively. ImageNet hyperparameters are also included in the Appendix in Table \ref{tab:manual_config}.

\section{Confusion Matrix for Uniform Asymmetric Noise}
\label{appendix:asymsetup}

The confusion matrix for uniform asymmetric noise is given in Equation \ref{eqn:c_asym_rte}.

\begin{table*}
$~$
\begin{equation}
\label{eqn:c_asym_rte}
C = \begin{pmatrix} 
.0000 & .0396 & .2475 & .0594 & .0594 & .0396 & .0495 & .0693 & .2772 & .1584 \\
.1765 & .0000 & .0294 & .0000 & .0000 & .0000 & .0294 & .0000 & .1765 & .5882 \\ 
.1745 & .0000 & .0000 & .1544 & .1879 & .1074 & .2617 & .0872 & .0268 & .0000 \\ 
.0388 & .0116 & .1473 & .0000 & .1240 & .3682 & .1899 & .0853 & .0155 & .0194 \\ 
.0303 & .0000 & .2197 & .1667 & .0000 & .0606 & .2879 & .2121 & .0227 & .0000 \\ 
.0324 & .0000 & .1435 & .4676 & .1019 & .0000 & .1204 & .1157 & .0093 & .0093 \\ 
.0536 & .0179 & .3571 & .3036 & .1071 & .0714 & .0000 & .0536 & .0179 & .0179 \\ 
.0704 & .0000 & .0986 & .1268 & .3803 & .1831 & .0986 & .0000 & .0000 & .0423 \\ 
.4603 & .0952 & .0794 & .0476 & .0317 & .0000 & .0476 & .0317 & .0000 & .2063 \\ 
.1711 & .5132 & .0263 & .0526 & .0263 & .0132 & .0658 & .0395 & .0921 & .0000 
\end{pmatrix} 
\end{equation}
\end{table*}

\section{PBT Experiments}
\label{PBThyper}

\begin{figure*}[h]
  \centering
  \includegraphics[width=1.0\textwidth]{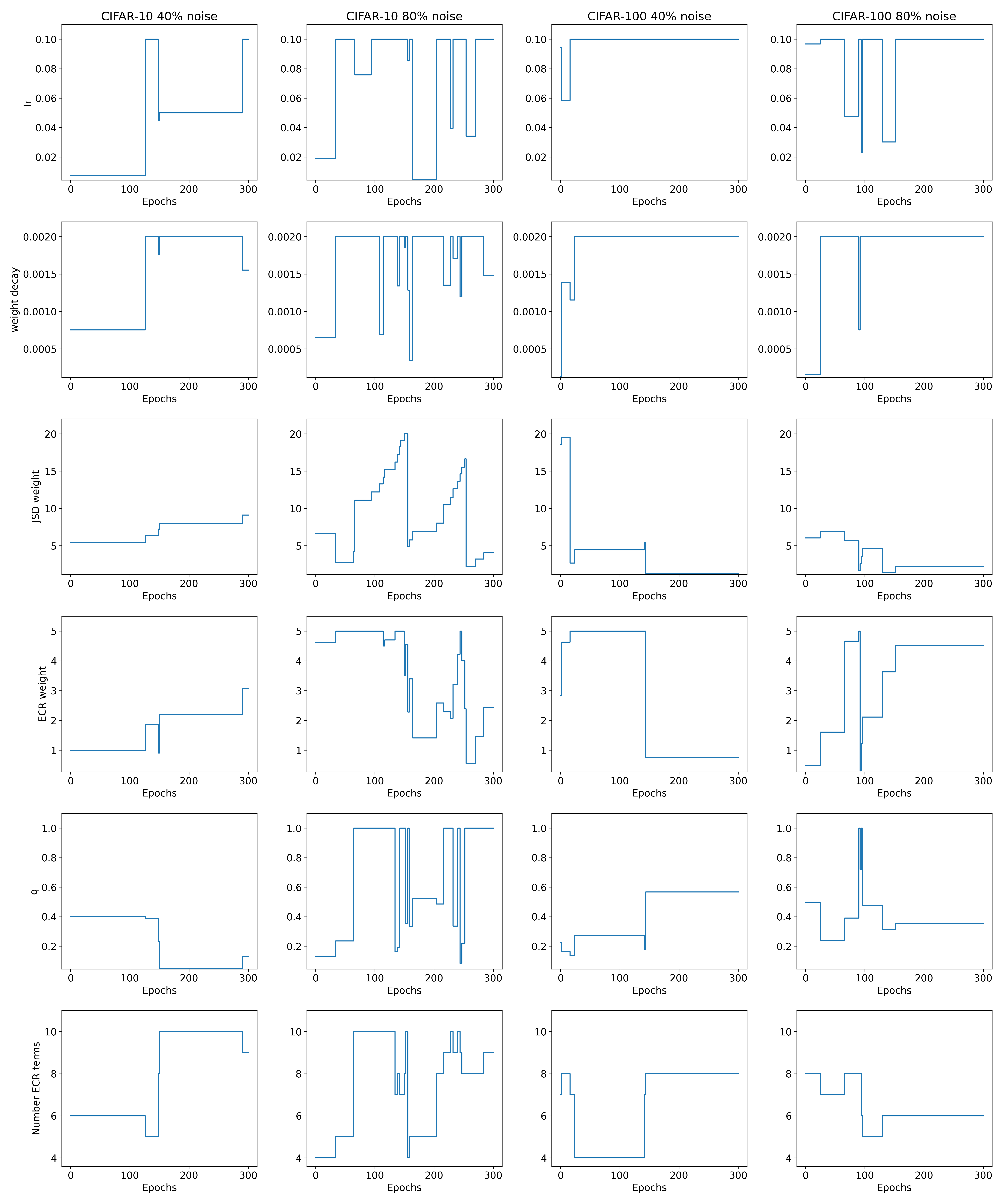}
  \caption{Parameter schedules for $lr$, $\text{weight decay}$, JSD weight $\lambda_{JSD}$, ECR weight $\lambda_{ECR}$, $q$ and $N^*$ for CIFAR-10 and CIFAR-100 with 40\% and 80\% uniform symmetric noise rates.}
  \label{pbt_all_schedules}
\end{figure*}

PBT sampling configurations are shown in Table \ref{table_pbtconfiguration}, and parameter schedules are shown in Figure \ref{pbt_all_schedules}.

\begin{table}
\centering
\caption{PBT sampling configuration for CIFAR-10 and CIFAR-100. We used a population size of 35, and permutation interval of 2 epochs. In the case a member inherits another checkpoint, each hyperparameter is resampled from its distribution with $p=0.25$ or is multiplied with $w \sim \text{Uniform}(0.8,1.2)$ within its parameter range with $p=0.75$. In the case of $N^*$, the previous/next hyperparameter from the ordered list is selected.}
\begin{tabular}{ll}
\toprule
Parameter    & Sample distribution \\
\midrule
Batch size        & 128                    \\
Dropout           & 0.01                    \\
$\beta$             & 0.9                    \\
$\alpha$            & 0.99                    \\
LR                & Uniform(0.00001, 0.1)                     \\
weight decay      & Uniform(0.00005, 0.002)                    \\
$q$               & Uniform(0.0, 1.0)                 \\
$\lambda_{\JSD}$    & Uniform(0.0, 20.0)                    \\
$\lambda_{\ECR}$    & Uniform(0.0, 5.0)                    \\
$N^*$             & Uniform\{3, ..., 10\}  \\
\bottomrule
\label{table_pbtconfiguration}
\end{tabular}
\end{table}

\section{Appendix: Uniform Asymmetric Noise on CIFAR-10}
\label{appendix:asym}

The matrix $C$ in Equation \ref{eqn:c_asym_patrini} defines the noise structure for uniform asymmetric noise on CIFAR-10 with following labels: AIRPLANE, AUTOMOBILE, BIRD, CAT, DEER, DOG, FROG, HORSE, SHIP, TRUCK.

Class distributions are shown in Table \ref{tab:c_asym_sample_60}.

\begin{table*}
\begin{equation}
\label{eqn:c_asym_patrini}
 C = \left( \begin{matrix} 0 & 0 & 0 & 0 & 0 & 0 & 0 & 0 & 0 & 0 \\ 0 & 0 & 0 & 0 & 0 & 0 & 0 & 0 & 0 & 0 \\ 1 & 0 & 0 & 0 & 0 & 0 & 0 & 0 & 0 & 0 \\ 0 & 0 & 0 & 0 & 0 & 1 & 0 & 0 & 0 & 0 \\ 0 & 0 & 0 & 0 & 0 & 0 & 0 & 1 & 0 & 0 \\ 0 & 0 & 0 & 1 & 0 & 0 & 0 & 0 & 0 & 0 \\ 0 & 0 & 0 & 0 & 0 & 0 & 0 & 0 & 0 & 0 \\ 0 & 0 & 0 & 0 & 0 & 0 & 0 & 0 & 0 & 0 \\ 0 & 0 & 0 & 0 & 0 & 0 & 0 & 0 & 0 & 0 \\ 0 & 1 & 0 & 0 & 0 & 0 & 0 & 0 & 0 & 0 \end{matrix} \right) 
\end{equation}
\end{table*}

\label{sec:c_asym_sample_60}
\begin{table*}[h]
\renewcommand{\arraystretch}{1.3}
\centering
  \caption{Overview of class distribution of total and correct labels after sampling noisy CIFAR-10 training labels with asymmetric noise defined by equation (\ref{eqn:c_asym_rte}) with a uniform 60\% noise ratio.\newline}
  
\begin{tabular}{lrrrr}
\toprule
          & \multicolumn{1}{l}{\# samples} & \multicolumn{1}{l}{\% samples} & \multicolumn{1}{l}{\# correct labels} & \multicolumn{1}{l}{\% correct labels} \\ \hline
\small{AIRPLANE}   & 5578                           & 11\%                           & 1958                                  & 35\%                                  \\
\small{AUTOMOBILE} & 4069                           & 8\%                            & 2003                                  & 49\%                                  \\
\small{BIRD}       & 6023                           & 12\%                           & 2017                                  & 33\%                                  \\
\small{CAT}        & 6205                           & 12\%                           & 2038                                  & 33\%                                  \\
\small{DEER}       & 5056                           & 10\%                           & 1986                                  & 39\%                                  \\
\small{DOG}        & 4480                           & 9\%                            & 1977                                  & 44\%                                  \\
\small{FROG}       & 5476                           & 11\%                           & 2019                                  & 37\%                                  \\
\small{HORSE}      & 4130                           & 8\%                            & 2028                                  & 49\%                                  \\
\small{SHIP}       & 3896                           & 8\%                            & 2024                                  & 52\%                                  \\
\small{TRUCK}      & 5087                           & 10\%                           & 1950                                  & 38\%                                  \\ \hline
TOTAL      & 50000                          & 100\%                          & 20000                                 & 40\%                                  \\ 
\bottomrule
\end{tabular}
\label{tab:c_asym_sample_60}
\end{table*}

\section{Appendix: Extended Data and Analysis}
\label{sec:ablation_more_data}

\begin{figure*}
  \centering
  \includegraphics[width=2.5in]{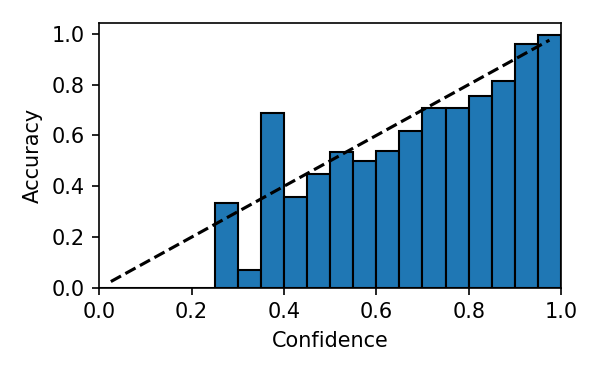}
  \includegraphics[width=2.5in]{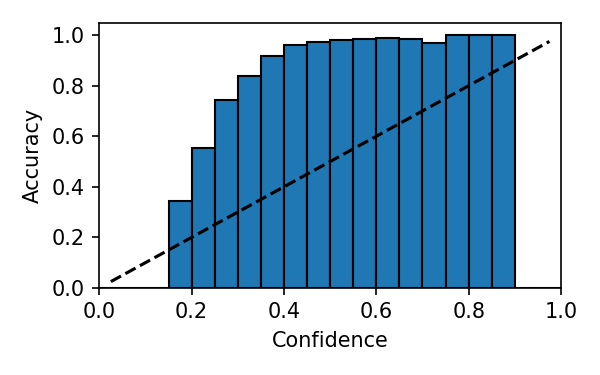}
  \caption{Reliability diagrams for RTE training models on CIFAR-10 with 40\% uniform label noise (left) and 80\% label noise (right). Perfectly calibrated models follow the black line, whereas over-confident models lie below and under-confident models lie above. This figure indicates our RTE trained model is well calibrated when trained with 40\% label noise, while (perhaps justifiably) conservative when trained with a more extreme level of 80\% label noise.}
  \label{fig:reliability-diagrams}
\end{figure*}

In Tables \ref{tab:fig1_left} and \ref{tab:fig1_right} we include test accuracy and mean corruption error on CIFAR-10 and CIFAR-10-C. In Figure \ref{fig:reliability-diagrams}, we include reliability diagrams using CIFAR-10.

\begin{table*}[h]
    \renewcommand{\arraystretch}{1.3}
    \centering
    \caption{RTE test accuracy and mean corruption error (mCE) on CIFAR-10 and CIFAR-10-C, respectively.  In this experiment, fixed batch size of $bs=128$ is used and the number of ECR terms, $N^*$ is varied.  Training configuration of these data is described in section \ref{sec:expt:us}.  Test accuracy is presented in Figure \ref{fig:ecr_term_batch_size} (left).}
    \begin{tabular}{lcccccccc}
        \toprule
        &\multicolumn{8}{l}{CIFAR-10} \\
        &\multicolumn{8}{l}{Fixed batch-size: 128} \\
        &\multicolumn{8}{l}{Uniform Symmetric Noise: 80\%} \\
        \hline
        &\multicolumn{8}{l}{Vary the number of ECR terms: $N^*$} \\
         & 1 & 2 & 3 & 4 & 5 & 6 & 7 & 8 \\
         \hline
         $\uparrow$ Test Acc & 91.51 & 91.90 & 92.57 & 92.65 & 92.77  & 93.14 & 93.09 & 93.21 \\
         $\downarrow$ mCE & 15.32 & 14.87 & 13.74 & 13.90 & 13.84 & 13.48 & 13.67 & 13.66 \\
         \bottomrule
    \end{tabular}
    \label{tab:fig1_left}
\end{table*}

\begin{table*}
    \renewcommand{\arraystretch}{1.3}
    \centering
    \caption{RTE test accuracy and mean corruption error (mCE) on CIFAR-10 and CIFAR-10-C, respectively.  In this experiment a single consistency loss term is used and vary the batch size of that term. This experiment with varying batch size is analogous to a more traditional semi-supervised approach where large batch size is used for unsupervised loss terms.  Training configuration for these data is described in section \ref{sec:expt:us}.  Test accuracy is presented in Figure \ref{fig:ecr_term_batch_size} (right).}
    \begin{tabular}{lcccccc}
        \\
        \toprule
        &\multicolumn{6}{l}{CIFAR-10} \\
        &\multicolumn{6}{l}{Fixed ECR terms: $N^*=1$} \\
        &\multicolumn{6}{l}{Uniform Symmetric Noise: 80\%} \\
        \hline
        &\multicolumn{6}{l}{Vary the batch size:} \\
         & 32 & 64 & 128 & 256 & 512 & 1024  \\
         \hline
         $\uparrow$ Test Acc & 86.54 & 88.95 & 90.32 & 88.46 & 85.87 & 78.13 \\
         $\downarrow$ mCE & 19.77 & 17.78 & 16.41 & 18.20 & 20.42 & 28.57 \\
         \bottomrule
    \end{tabular}
    \label{tab:fig1_right}
\end{table*}

\section{Appendix: Compute Resources}

We used an internal cluster of NVIDIA V100s for all experiments. We estimate that all experiments across all datasets (CIFAR-10, CIFAR-100, ImageNet, WebVision, and Food-101N) required approximately 2,000 GPU hours.

\end{document}